\documentclass[letterpaper, 10 pt, conference]{ieeeconf}

\IEEEoverridecommandlockouts
\usepackage{amsmath} % assumes amsmath package installed
\usepackage{amssymb}  % assumes amsmath package installed
\usepackage[protrusion=false]{microtype}
\usepackage{tensor}
\usepackage{mathtools}
\usepackage{makecell}
\usepackage[colorinlistoftodos,prependcaption]{todonotes}
\usepackage{siunitx}
\usepackage{comment}
\usepackage{xcolor} % for textcolor
\usepackage[normalem]{ulem}
\usepackage{esvect}
\usepackage{float}
\usepackage{lipsum}
\usepackage{multicol}
\usepackage{multirow}
\usepackage{flushend}
\usepackage[linesnumbered,ruled,vlined]{algorithm2e}

\usepackage{booktabs}
\usepackage{todonotes}

\title{\LARGE \bf
	Observability-aware online multi-lidar extrinsic calibration}

\author{Sandipan Das$^{1,2}$, Ludvig af Klinteberg$^2$, Maurice Fallon$^3$, Saikat Chatterjee$^{1}$
	\thanks{$^1$ KTH EECS, Sweden. \texttt{\{sandipan,sach\}@kth.se}\newline%
			$^2$ Autonomous Transport Solutions Research, Scania CV AB, Sweden. \texttt{\{sandipan.das, ludvig.af.klinteberg\}@scania.com}\newline%
			$^3$ Oxford Robotics Institute, UK. \texttt{mfallon@robots.ox.ac.uk}}}

\usepackage[linesnumbered,ruled,vlined]{algorithm2e}

\usepackage{multirow}
\usepackage[normalem]{ulem}
\usepackage{xargs} 

\newcommand{\hide}[1]{}
\newcommand{\Figure}{Fig.~}
\newcommand{\Equation}{Eq.~}

\newcommand{\bdmath}{\begin{dmath}}
\newcommand{\edmath}{\end{dmath}}
\newcommand{\beq}{\begin{equation}}
\newcommand{\eeq}{\end{equation}}
\newcommand{\bdm}{\begin{displaymath}}
\newcommand{\edm}{\end{displaymath}}
\newcommand{\bea}{\begin{eqnarray}}
\newcommand{\eea}{\end{eqnarray}}
\newcommand{\beal}{\beq \begin{array}{ll}}
\newcommand{\eeal}{\end{array} \eeq}
\newcommand{\beas}{\begin{eqnarray*}}
\newcommand{\eeas}{\end{eqnarray*}}
\newcommand{\ba}{\begin{array}}
\newcommand{\ea}{\end{array}}
\newcommand{\bit}{\begin{itemize}}
\newcommand{\eit}{\end{itemize}}
\newcommand{\ben}{\begin{enumerate}}
\newcommand{\een}{\end{enumerate}}

\newcommand{\SEthree}{\ensuremath{\mathrm{SE}(3)}\xspace}

\newcommand{\T}{\mathbf{T}}
\newcommand{\R}{\mathbf{R}}

\newcommand{\Identity}{\mathbf{I}}

\newcommand{\tran}{\mathbf{t}}

\newcommand{\eq}{Eq.}

\newcommand{\World}{\mathtt{W}}
\newcommand{\Imu}{\mathtt{I}}
\newcommand{\Camera}{\mathtt{C}}
\newcommand{\Lidar}{\mathtt{L}}
\newcommand{\GNSS}{\mathtt{G}}

\newcommand{\imu}{\mathtt{{I}}}

\newcommand{\Base}{\mathtt{{B}}}

\SetCommentSty{algo}
\SetKwInput{KwInput}{Input} % Set the Input
\SetKwInput{KwOutput}{Output} % set the Output

\newcommand{\etalcite}[2]{#1~et~al.~\cite{#2}}

\DeclareMathOperator*{\argmin}{arg\,min}

\newcommand\cancel{\bgroup\markoverwith{\textcolor{red}{\rule[0.5ex]{2pt}{0.4pt}}}\ULon}

\makeatletter
\let\NAT@parse\undefined
\makeatother
\usepackage{hyperref}
\begin{document}
	
	\maketitle
	\thispagestyle{empty}
	\pagestyle{empty}
	
        \begin{abstract}
	Accurate and robust extrinsic calibration is necessary for deploying autonomous systems which need multiple sensors for perception. In this paper, we present a robust system for real-time extrinsic calibration of multiple lidars in vehicle base frame without the need for any fiducial markers or features. We base our approach on matching absolute GNSS and estimated lidar poses in real-time. Comparing rotation components allows us to improve the robustness of the solution than traditional least-square approach comparing translation components only. Additionally, instead of comparing all corresponding poses, we select poses comprising maximum mutual information based on our novel observability criteria. This allows us to identify a subset of the poses helpful for real-time calibration. We also provide stopping criteria for ensuring calibration completion. To validate our approach extensive tests were carried out on data collected using Scania test vehicles (7 sequences for a total of $\approx$ 6.5 Km). The results presented in this paper show that our approach is able to accurately determine the extrinsic calibration for various combinations of sensor setups.
      \end{abstract}

	% Keep Consistent Tenses:
	%in background/lit review: past
	%for experiments carried out: past
	%for results being evaluated currenty: present

	\section{Introduction}
	\label{sec:introduction}
	
	Autonomous vehicles require multiple sensors to create $360^{\circ}$ sensing coverage for navigating safely and redundancy. Thus it is important to determine the accurate mounting position of the sensors in the vehicle which is referred to as extrinsic calibration. It is not possible to fuse the sensing data in a common reference frame
	without correct extrinsic calibration parameters. In our work we focused on extrinsic calibration of multi-lidar systems.
	
	Extrinsic calibration can be performed in 2 ways: Offline and Online.
	Offline procedure refers to matching features between the lidars by placing fiducial markers in their common field-of-view (FoV) of both the sensors \cite{He2013, Choi2016}. Online calibration is performed  by matching the estimated states of the sensor in its corresponding sensor frame \cite{brookshire2013extrinsic, heng2013camodocal, furgale2013unified, huang2018geometric, jiao2021robust, lv2022observability}. Its desirable to perform online calibration since the sensors may not have common FoV and at the same time it reduces engineering effort compared to offline process.
	
	Online calibration methods rely on motion of the mobile platform. At the same time it is well known \cite{kalibr_meye_2013,hausman2017observability, schneider2019observability}, all motion segments do not provide enough information necessary for extrinsic calibration.
	Hence, identifying the degenerate motion segments help to discard poses which are not necessary for extrinsic calibration computation.
	
	We provide an observability-aware online calibration framework which can be run in the vehicle for calibrating multiple lidars in real-time. We also provide an analytical study on the effect noisy state estimates in the online calibration process.
	
	\begin{figure}
		\centering
		\includegraphics[width=1\columnwidth]{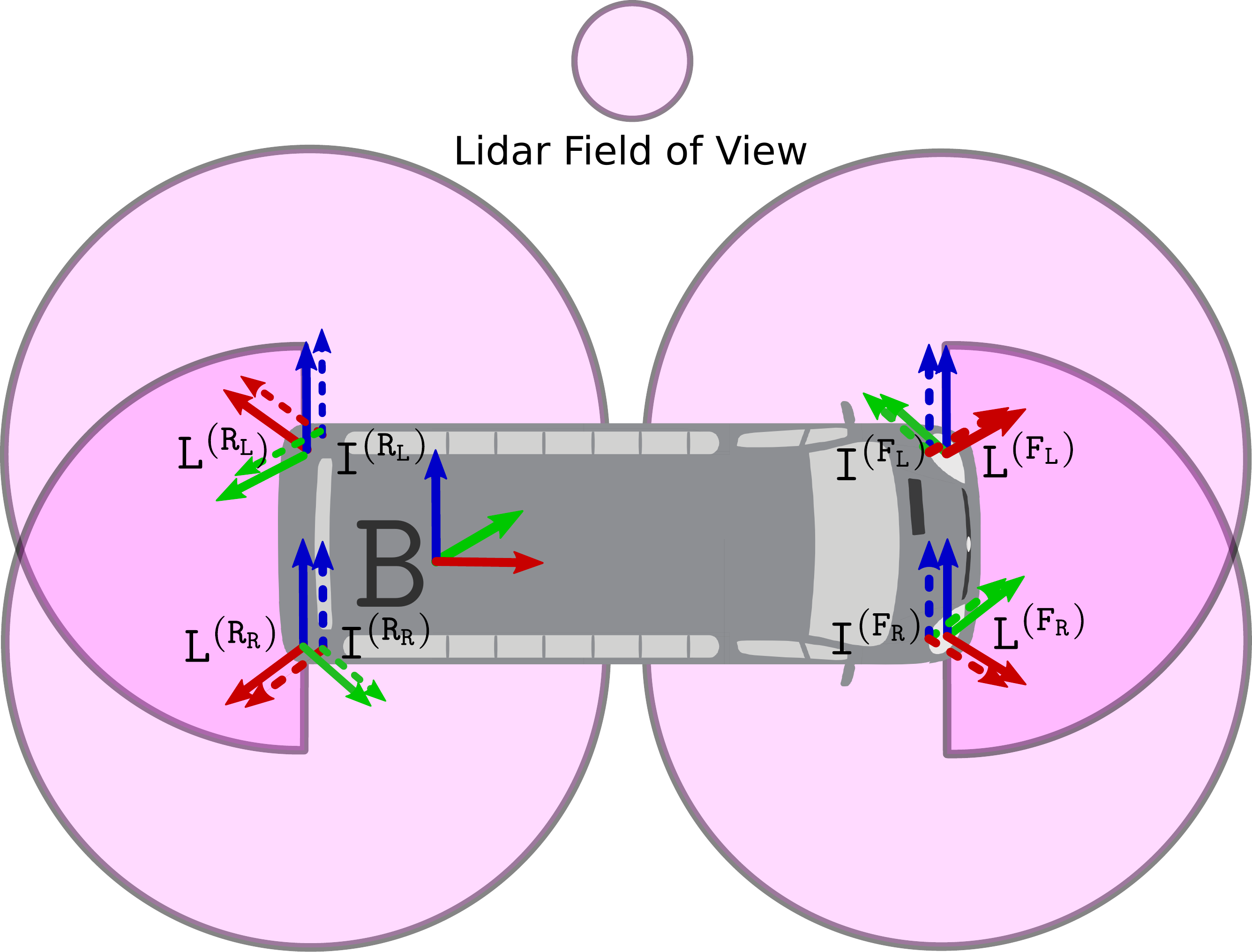}
		%\vspace{-5mm}
		\caption{Illustration of the four lidars with their embedded IMUs positioned
			around one data collection vehicle, when the sensors are calibrated. The vehicle base frame $\Base$, is located
			at the center of the rear axle. The sensor frames of the lidars are:
			$\Lidar^{(\mathtt{F}_\mathtt{L})}$, $\Lidar^{(\mathtt{F}_\mathtt{R})}$, $\Lidar^{(\mathtt{R}_\mathtt{R})}$ and
			$\Lidar^{(\mathtt{F}_\mathtt{L})}$, represented in solid lines; whereas, the sensor frames of the IMUs are:
			$\Imu^{(\mathtt{F}_\mathtt{L})}$, $\Imu^{(\mathtt{F}_\mathtt{R})}$, $\Imu^{(\mathtt{R}_\mathtt{R})}$ and $\Imu^{(\mathtt{R}_\mathtt{L})}$, represented in dashed lines.
			${(\mathtt{F}_\mathtt{L})}$: front-left, ${(\mathtt{F}_\mathtt{R})}$: front-right, ${(\mathtt{R}_\mathtt{R})}$: rear-right and
			${(\mathtt{R}_\mathtt{L})}$: rear-left.} 
		\label{fig:scania-bevda}
		\vspace{-5mm}
	\end{figure}
	
	\subsection{Motivation}
	The mobile platforms used in our experiments have multitude of sensors as shown in \Figure\ref{fig:scania-bevda}. Moreover, before any autonomous run it is necessary to ensure that the sensors are properly calibrated. So we need to perform real-time calibration of all the sensors \textit{wrt} to a common reference frame, usually located in center of the rear-axle. We also need to identify motion segments which provide enough excitation for different degrees of freedom required for calibration. Finally we need a mechanism to understand when the online calibration performance is satisfactory. 
	
	To achieve this, we calibrate all the lidars \textit{wrt} a GNSS system which reports pose of the common reference frame in global frame. The GNSS system additionally reports the inertial measurements of the vehicle. All the lidar sensors used in our experiments also have an embedded IMU. Hence, for observability analysis we compare only the raw angular velocity signals between the GNSS system and the embedded IMU within the lidar(s) unlike comparing estimated states which inherently has process noise. Finally, we introduce a cost function to stop the calibration process when the results are satisfactory.

	\subsection{Contribution}
	\label{sec:contribution}
	
	Our work is motivated by the broad literature in motion based calibration. Our proposed contributions are:
	\begin{itemize}
		\item An extrinsic calibration algorithm to calibrate multiple lidars in real-time using GNSS reference poses with a calibration completion criteria.
		\item A novel approach for observability analysis based on comparing raw angular velocity signals.
		\item We also provide our analytical study on sensitivity of the estimated states used in our online calibration algorithm.
		\item Verification of our method based on data collected from Scania autonomous vehicles with the sensor setup shown in Fig. \ref{fig:coordinate-frames},
		with FoV schematics similar to Fig. \ref{fig:scania-bevda}.
	\end{itemize}

	\section{Related Work}
	\label{sec:related-work}
	
	Online extrinsic calibration based on different strategies is an widely studied area. In our discussion we briefly review the relevant literature and motivate our choice of method adapted in our work.
	
	\subsection{Motion-based methods for online extrinsic calibration}
	Online extrinsic calibration based on the Hand-Eye method \cite{hand-eye_tsai_1989,horaud1995hand} is an extensively studied topic. 
	Initial research on motion-based calibration aligned a gripper (hand) and a
	camera (eye) by estimating the motion of the gripper along with the camera and
	constraining their poses with a fixed rigid body transformation, this approach is called Hand-Eye calibration and has been reviewed by
	\etalcite{Shah}{shah2012overview}. Multi-camera extrinsic calibration with
	the Hand-Eye method have been explored in works like Camodocal
	\cite{heng2013camodocal} and recently extended to multi-lidar extrinsic calibration \cite{jiao2019automatic, multilidar2021das}.

	Instead of performing Hand-Eye
	calibration, in Kalibr \cite{kalibr_meye_2013} the authors automatically
	detected sets of measurements from which they could identify an observable parameter space
	and then performed a maximum likelihood estimate (MLE) by minimizing the errors
	between landmark observations and their known correspondences. They also
	discarded parameter updates for numerically unobservant directions and degenerate
	scenarios. \etalcite{Furgale}{furgale2013unified} further generalized Kalibr for
	spatio-temporal calibration. We build upon the philosophy of Kalibr and
	minimize the error between the poses estimated by the lidar(s) and the GNSS
	in a batch optimization process to recover the extrinsic calibration parameters.
	
	\subsubsection{Types of motion models}
	\etalcite{Huang}{huang2018geometric} proposed a framework to evaluate different motion based strategies with their work on noise sensitivity analysis. They categorized the motion-based methods as:
	\begin{itemize}
	\item Model A: Comparing absolute poses in a common frame.
	\item Model B: Comparing absolute poses in separate frames.
	\item Model C: Comparing relative poses -- Hand-Eye method.
	\end{itemize}
	They show that Model A performs well for all the relative angles between the pose pairs and has minimum number of degeneration zones. Hence, we base our cost function to recover transformation based on Model A.  
	
	\subsubsection{Types of cost functions}
	We can formulate the cost function in 2 different approaches:
	\begin{itemize}
		\item Minimize the absolute translation in a least-square approach \cite{Kabsch1978, horn1987closed}.
		\item Assume additive Gaussian noise in the estimated poses and minimize the weighted noise \cite{huang2017extrinsic, jiao2021robust} component.
	\end{itemize}
	In our work we formulate our cost function as a least-square problem.
	Unlike comparing only the translation component for comparing the poses we also compare the rotation components using the Q-method by Davenport \cite{davenport1968vector}. Additionally, we analytically show that optimization problem remains invariant under certain noisy conditions. We also base our novel observability criteria based on comparing the raw angular velocity signals and stop the calibration process when satisfactory performance is reached.

	%Instead, we based our MIMU calibration
	%on the physical principle that angular velocity of a rigid body about any %point
	%of that body is same. We compare the raw angular velocities in between the
	%IMU(s) and GNSS using MLE and recover the extrinsics following the principle of
	%signal-to-signal match \cite{ilewicz2013direct} in a batch optimization
	%framework.
	
	\subsection{Lidar odometry}
	\label{sec:loam}

	There have been vast array of works on lidar odometry
	after the LOAM paper by \etalcite{Zhang}{Zhang2017}. Often geometric features are extracted
	from the point clouds and tracked between frames for computational efficiency of various state estimation algorithms
	\cite{wisth2020vilens, shan-legoloam, tixiao2020lio-sam}. However, this principle doesn't work for featureless
	environments. Hence, instead of feature extraction, the whole point cloud is often
	processed, which is known as direct estimation, shown in works like Fast-LIO2 \cite{xu2022fast} and Faster-LIO
	\cite{bai2022faster}. We adapted Fast-LIO2 for estimating lidar odometry in our online calibration algorithm.

	\section{Problem Statement}
	\label{sec:problem-statement}

	\subsection{Sensor platform and reference frames}
	\label{sec:sensor-platform}
	\begin{figure}
		\centering
		\vspace{2mm}
		\includegraphics[width=1\columnwidth]{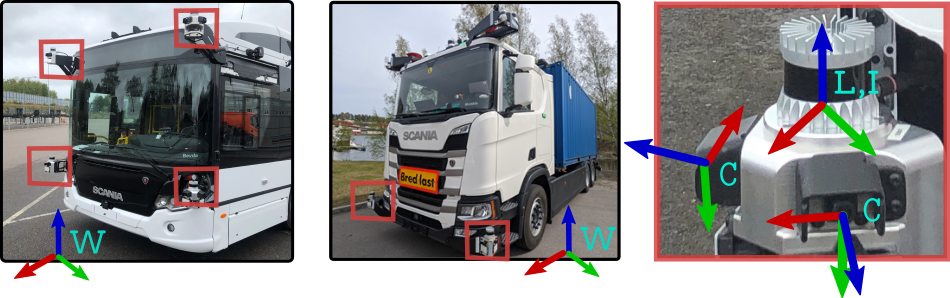} 
		\vspace{-8mm}
		\caption{Reference frames conventions for our bus and truck platforms. In both cases, the world frame
			$\World$ is a fixed frame, while the base frame $\Base$, as shown in \Figure
			\ref{fig:scania-bevda}, is located at the rear axle center of the vehicle.
			Each
			sensor unit contains the two optical frames $\Camera$, an IMU frame, $\Imu$,
			and
			lidar frame $\Lidar$.} %\sandipan{I think we can improve this figure by
		%adding
		%a schematic of frames, in ros we can model the robot by urdf model}
		\label{fig:coordinate-frames}
		\vspace{-5mm}
	\end{figure}
	%\mfallon{should you show a slight difference between lidar and inertial frames - given you are working on calibration you should be precise}

	The sensor configuration used is as shown in \Figure
	\ref{fig:coordinate-frames} for a bus and truck. We also show illustrations of the fields-of-view of the sensors for the bus in \Figure\ref{fig:scania-bevda}. Each of the sensor housings contains a lidar with an integrated IMU and two cameras. Please note the cameras are not used for this work.
	
	First we will describe the necessary notation and the reference frames used in our system. The vehicle base frame, $\Base$ is located on the center of the rear-axle of the vehicle. Sensor readings from GNSS, lidars and IMUs are represented in their respective sensor frames as $\GNSS$, $\Lidar^{(k)}$ and $\Imu^{(k)}$ respectively. Here, $k \in [\mathtt{F}_\mathtt{L}, \mathtt{F}_\mathtt{R}, \mathtt{R}_\mathtt{L}, \mathtt{R}_\mathtt{R}]$ denotes the location of the sensor in the vehicle corresponding to front-left, front-right, rear-left and rear-right respectively. The GNSS measurements are reported in world fixed frame, $\World$ and transformed to $\Base$ frame by performing a calibration routine outside the scope of this work.
	In our discussions the transformation matrix is denoted as, $\T =
	\left[\begin{array}{ll}\mathbf{R}_{3\times3} & \mathbf{t}_{3\times1} \\
		\mathbf{0}^{\top} & 1\end{array}\right]\in \SEthree$ and
	$\R\R^T=\mathbf{I}_{3\times3}$, since the rotation matrix is orthogonal. 
	
	\subsection{Problem formulation}
	Our primary goal is to estimate the extrinsic calibration of multiple
	lidar sensors \textit{wrt} the base frame and to recover
	$\T_{\Base\Lidar^{(k)}}$ in real-time without the need for any fiducial markers. Since, our method is based on a motion-based approach, we first estimate the pose of the lidar sensor, $\T_{\mathtt{L}_i^{(k)}}$, at time $t_i$ in its corresponding sensor frame, $\Lidar^{(k)}$, prior to the online calibration. We also study the effect of process noise in our algorithm and identify the states which provide maximum excitation for different degrees of freedom, based on which online extrinsic calibration is performed.
	
	\section{Methodology}
	\label{sec:methodology}
	
	\subsection{Initialization}	
	All the GNSS poses in base frame, $\Base$ are normalized \textit{wrt} to the initial pose, $\tensor[_\World]{\T}{_{\mathtt{{\World\GNSS}}}}$. The GNSS
	unit also provides IMU measurements in the $\GNSS$ frame using its internally
	embedded IMU. These measurement are then transformed to base frame, $\Base$. For gravity alignment we use the equations 25 and 26 from \cite{pedley2013tilt} to estimate roll and pitch after collecting IMU data when the vehicle is static for few seconds according to the GNSS estimate.

	%	\begin{align}
	%		\operatorname{\widehat{Roll}} &=
	%\tan^{-1}\left(\frac{\sum_{i=1}^{t}{\mathbf{f}}_{y_i}}{\sum_{i=1}^{t}\mathbf{f}_{z_i}}\right)
	%\\
	%		\operatorname{\widehat{Pitch}} &=
	%\tan^{-1}\left(\frac{-{\sum_{i=1}^{t}\mathbf{f}}_{x_i}}{\sqrt{\sum_{i=1}^{t}(\mathbf{f}_{y_i}^2
	%+ \mathbf{f}_{z_i}^2)}}\right)
	%	\end{align}
	%where, $\mathbf{f}$ is the linear acceleration from the internally embedded IMU
	%from GNSS system. 

	Note that the noise processes and starting bias estimates for the embedded IMU sensor within the lidars and the GNSS unit 
	were characterized in advance by estimating the Allan Variance 
	\cite{allan1966statistics} parameters using logs collected while the vehicle was stationary.
	
	%We compute average of the IMU signals $n$ times in a
	%standstill position for $t$ seconds. The formulation of Allan variance of a
	%signal $\mathbf{s}$, is given as,
	%	\begin{align}
	%		\operatorname{Allan Variance}(t) &=
	%\frac{1}{2(n-1)}\sum_{i=1}^{n-1}\left[\mathbf{s}(t)_{i+1} -
	%\mathbf{s}(t)_{i}\right]^2 \label{eq:allan-var}\\
	%		\operatorname{Allan Deviation}(t) &= \sqrt{\operatorname{Allan Variance}(t)}
	%		\label{eq:allan-std}
	%	\end{align}
	
	% A very interesting read on time-keeping:
	%http://www.allanstime.com/AllanVariance/
	
	%Allan variance: https://www.cl.cam.ac.uk/techreports/UCAM-CL-TR-696.pdf
	
	\subsection{Lidar-IMU calibration}	
	%Each lidar has an integrated IMU. 
	The integrated IMU within the lidar is utilized to rectify motion-induced lidar distortion as well as to pre-compute the initial guess for scan matching within the lidar odometry. Hence, it is important to calibrate the lidar and its corresponding integrated IMU. If the sensor supplier provided extrinsics are not available, the calibration can be formulated as a Hand-Eye problem to estimate the relative IMU and lidar poses separately. We integrate the IMU measurements between estimated lidar poses using IMU preintegration \cite{Forster2017}. The state propagation equations are based on the IMU data $[\boldsymbol{\omega}: \text{angular velocity}, \mathbf{f}: \text{linear acceleration}]$ between two timestamps [$t_i$, $t_j$] for all the $\Delta t$ (inverse of sensor frequency) intervals in the \SEthree manifold is derived in \cite{Forster2017} as:
	\begin{equation}
		\begin{aligned} 
			\Delta \mathtt{R}_{i j} & =\prod_{k=i}^{j-1} \operatorname{Exp}\left(\boldsymbol{\omega}_{\Imu_k} \Delta t\right), \\ 
			\Delta \mathbf{v}_{i j} & =\sum_{k=i}^{j-1} \Delta \mathrm{R}_{i k}\mathbf{f}_{\Imu_k} \Delta t, \\ 
			\Delta \mathbf{p}_{i j} & =\sum_{k=i}^{j-1}\left[\Delta \mathbf{v}_{i k} \Delta t+\frac{1}{2} \Delta \mathrm{R}_{i k}\mathbf{f}_{\Imu_k} \Delta t^{2}\right],
			\label{eq:imu-preintegration}
		\end{aligned}
              \end{equation}
    where, $\Delta \mathtt{R}_{i j}, \Delta \mathbf{v}_{i j}, \Delta \mathbf{p}_{i j}$ represent the relative rotation, velocity and position respectively.
	Based on the computed lidar \ref{sec:loam} and IMU relative poses (\eq~\ref{eq:imu-preintegration}) we formulate a Hand-Eye problem as,
	\begin{equation}
		{\T}{_{\Lidar_{i}\Lidar_j}}{\T}{_{\Lidar\Imu}} = {\T}{_{\Lidar\Imu}}{\T}{_{{\Imu_{i}\Imu_j}}}.
		\label{eq:hand-eye-li}
	\end{equation}
	Since, we use the relative transforms from the IMU propagation,  $\mathbf{R}{_{\Lidar\Imu}^{\star}}$ is determined by equating the rotation parts in \eq~\ref{eq:hand-eye-li} in quaternion form, similar to \cite{lv2022observability} as, 
	%\left[q_s, \mathbf{q}_v\right]

    %\mfallon{you could be more consistent on the following: A `sensor pose' is relative to the fixed real-world. `relative poses' are output from odometry - and are relative to the previous pose. a transformation is the relative pose between two fixed frames: imu and lidar or camera.}

	\begin{equation}
		\begin{aligned} 
			& &{\mathbf{q}}{_{\Lidar_{i}\Lidar_j}}\otimes{\mathbf{q}}{_{\Lidar\Imu}} = {\mathbf{q}}{_{\Lidar\Imu}}\otimes{\mathbf{q}}_{\Imu_{i}\Imu_j}\\
			& \implies & \underbrace{\left[\mathcal{L}(\mathbf{q}{_{\Lidar_{i}\Lidar_j}}) - \mathcal{R}(\mathbf{q}_{\Imu_{i}\Imu_j})\right]}_{\mathbf{A}_k} {\mathbf{q}}{_{\Lidar\Imu}} = 0, \label{eq:rot_lidar_imu}\\
		\end{aligned}
	\end{equation}
	where, the matrices, $\mathcal{L}(\mathbf{q})$ and $\mathcal{R}(\mathbf{q})$ are defined as,
	\begin{equation}
		\begin{aligned} 
			\mathcal{L}(\mathbf{q}) &=\left[\begin{array}{cc}q_{s} \mathbf{I}_{3}-\left[\mathbf{q}_{v}\right]_{\times} & \mathbf{q}_{v} \\ -\mathbf{q}_{v}^{T} & q_{s}\end{array}\right] \\
			\mathcal{R}(\mathbf{q}) &=\left[\begin{array}{cc}q_{s} \mathbf{I}_{3}+\left[\mathbf{q}_{v}\right]_{\times} & \mathbf{q}_{v} \\ -\mathbf{q}_{v}^{T} & q_{s}\end{array}\right], \nonumber
		\end{aligned}
	\end{equation}
	and, $\mathbf{q}$ is a quaternion represented as, 
	\begin{align}
		\mathbf{q} = \underbrace{q_s}_{\text {Scalar component}} +
		\underbrace{q_x\hat{i} + q_y\hat{j} + q_z\hat{k}}_{\text {Vector component}} =
		\left[q_s, \mathbf{q}_v\right]. 
	\end{align}

	For a set of relative poses we can form a system of linear equations of the form, $\mathbf{Ax} = \mathbf{0}$, where $\mathbf{A} = [\mathbf{A}_0^T \ \mathbf{A}_1^T \ \dots \ \mathbf{A}_n^T]^T$. ${\mathbf{q}}^{\star}_{\Lidar\Imu}$ lies in the nullspace of $\mathbf{A}$, which corresponds to the right unit singular vector with the smallest singular value of $\mathbf{A}$.
        
	After finding $\mathbf{q}{_{\Lidar\Imu}^{\star}}$, for the translation component we get,
	\begin{equation}
		\underbrace{\left(\mathbf{R}_{{\Lidar_{i}}{\Lidar_j}} - \mathbf{I}\right)}_{\mathbf{A}_k}\mathbf{t}{_{\Lidar\Imu}} \stackrel{\eq~\ref{eq:hand-eye-li}} {=} \underbrace{\left(\mathbf{R}{_{\Lidar\Imu}^{\star}}\mathbf{t}_{{\Imu_{i}}{\Imu_j}} -
			\mathbf{t}_{{\Lidar_{i}}{\Lidar_j}} \right)}_{\mathbf{B}_k},
		\label{eq:hand-eye-t-li}
              \end{equation}
	which is a system of linear equations of the form, $\mathbf{A}{\mathbf{x}} = \mathbf{B}$, where, $\mathbf{A} = [\mathbf{A}_0^T \ \mathbf{A}_1^T \ \dots \ \mathbf{A}^T_
	n]^T$, $\mathbf{B}^T = [\mathbf{B}_0^T \ \mathbf{B}_1^T \ \dots \ \mathbf{B}_
	n^T]^T$ and $\mathbf{x} = \mathbf{t}_{\Lidar\Imu}$.
	This is an over-determined system which can be solved using a least-square approach as,
	\begin{equation}
		\mathbf{x}^\star = \argmin_{\mathbf{x}}\| \mathbf{A}\mathbf{x} - \mathbf{B}\|^2_{\boldsymbol{\Sigma}},
		\label{eq:unbounded}
	\end{equation}
	where, ${\boldsymbol{\Sigma}}$ denotes the covariance of the residual. For these problems, the general strategy \cite{nocedal1999numerical} is to iteratively approximate the original problem by linearizing as, $F(\mathbf{x}+\Delta \mathbf{x}) \approx F(\mathbf{x})+\mathbf{J}(\mathbf{x}) \Delta \mathbf{x}$, where, $J$ being the Jacobian of $F(\mathbf{x})$. Thus, $\hat{\mathbf{x}}$ is updated in the current iteration as,
	$\hat{\mathbf{x}} \leftarrow \hat{\mathbf{x}} \boxplus \mathbf{\Delta x}$, where $\boxplus$ is an additional operator on the manifold and the problem then becomes,
	\begin{equation}
		\begin{aligned}
			&\mathbf{\Delta x}^\star = \argmin_{\mathbf{\Delta x}}\frac{1}{2}\| (\mathbf{A}\hat{\mathbf{x}} - \mathbf{B}) + \mathbf{A}^T\mathbf{\Delta x}\|^2_{\boldsymbol{\Sigma}}.
			\label{eq:unbounded_expanded}
		\end{aligned}
	\end{equation}
	The optimal solution is given by,
	\begin{equation}
		\underbrace{\left( \mathbf{J}^{\top} \boldsymbol{\Sigma}^{-1}\mathbf{J} \right)}_{\text{Fisher information matrix}}\mathbf{\Delta x}=-\mathbf{J}^{\top} \boldsymbol{\Sigma}^{-1} (\mathbf{A}\hat{\mathbf{x}} - \mathbf{B}).
		\label{eq:bvls_iterative_solution}
	\end{equation}
	%We used a bounded variable least-square solver \cite{Agarwal_Ceres_Solver_2022} with trust region approach and updated the parameters only if it improves the cost function, as defined in \eq~\ref{eq:error_term}.
	In practice, since we can take a reliable initial estimate for the translation component from the vehicle's CAD model, we search for $\mathbf{x}^\star$ only in a local neighborhood within reasonable bounds. Thus, we solve a bounded variable least squares problem as,
	\begin{equation}
		\mathbf{x}^\star = \argmin_{\mathbf{L} \le \mathbf{x} \le \mathbf{U}}\| \mathbf{A}\mathbf{x} - \mathbf{B}\|^2_{\boldsymbol{\Sigma}},
		\label{eq:bvls}
	\end{equation}
	where, $\mathbf{L}$ and $\mathbf{U}$ are the lower and upper bounds of $\mathbf{x}$ respectively. Thus the solution space in \eq~\ref{eq:unbounded_expanded} is modified with an additional constraint as, 
	\begin{equation}
		\begin{aligned}
			&\mathbf{\Delta x}^\star = \argmin_{\mathbf{\Delta x}}\frac{1}{2}\| (\mathbf{A}\hat{\mathbf{x}} - \mathbf{B}) + \mathbf{A}^T\mathbf{\Delta x}\|^2_{\boldsymbol{\Sigma}},\\
			& \textit{s.t.} \ \mathbf{L} \le \hat{\mathbf{x}} + \mathbf{\Delta x} \le \mathbf{U}.
			\label{eq:bvls_expanded}
		\end{aligned}
	\end{equation} 
	
	\subsection{Lidar calibration in base frame}
	Let, $\mathbf{T}_{\Base_i}$ be the pose of the base frame obtained using the GNSS
	measurements at timestamp $t_i$. And, $\T_{\Lidar_i^{(k)}}$ be the pose of the
	$k^{th}$ lidar in its corresponding sensor frame obtained from a state
	of the art state estimator \cite{xu2022fast} at timestamp $t_i$. Then our problem is to estimate the optimal
	transformation matrix between 2 pair of poses accumulated over $N$ timestamps:
	\begin{align}
		&\T_{\Base\Lidar^{(k)}}^{\star} =
		\left[\begin{array}{cc}\mathbf{R}^*_{3\times3} & \mathbf{t}^*_{3\times1} \\
			\mathbf{0}^{\top} & 1\end{array}\right] \nonumber\\
		&= \argmin_{\mathbf{T}_{\Base\Lidar^{(k)}}}
		\sum_{i=1}^{N}\|\mathbf{T}_{\Base\Lidar^{(k)}}\mathbf{T}_{\Base_i} -
		\mathbf{T}_{\Lidar_i^{(k)}}\|^2_F \label{eq:pose-alignment}\\
		&= \argmin_{\mathbf{R}_{\Base\Lidar^{(k)}}, \mathbf{t}_{\Base\Lidar^{(k)}}}\sum_{i=1}^{N} \resizebox{0.59\columnwidth}{!}{$\left\|\begin{array}{c|c}\mathbf{R}_{\Base\Lidar^{(k)}}\R_{\Base_i} - \R_{\Lidar_i} & \mathbf{R}_{\Base\Lidar^{(k)}}\tran_{\Base_i} + \tran_{\Base\Lidar^{(k)}} - \tran_{\Lidar_i} \label{eq:pose-alignment-expansion}\\ \\ \hline \\
			\mathbf{0}^{\top} & \mathbf{0}^{\top}\end{array}\right\|_F^2$}.
	\end{align}

	%\mfallon{equation 12 is a bit ugly.}
	%\sandipan{We are building upon this. It's important to show this.}
	
	\subsubsection{Initial calibration estimation}
	The initial extrinsics can be recovered by
	aligning the initial set of translation components, $\mathbf{t}_{\Base_i}$ and
	$\mathbf{t}_{\Lidar_i}$, $\forall i=1(1)N$, using a 3D-point cloud alignment
	method proposed by \etalcite{Kabsch}{Kabsch1978} or \etalcite{Horn}{horn1987closed}, by breaking the problem into rotation and translation parts. 	Note that we omit the $k^{th}$ superscript for brevity in the rest of our discussion.
	
	\subsubsection{Rotation estimation}
	For Frobenius norm,
	\begin{align}
		\|\mathbf{A} + \mathbf{B}\|_{\mathrm{F}}^{2}\stackrel{def} {=}\|\mathbf{A}\|_{\mathrm{F}}^{2}+\|\mathbf{B}\|_{\mathrm{F}}^{2}+2\langle \mathbf{A}, \mathbf{B}\rangle_{\mathrm{F}}
		\\ \text{where, } \langle\mathbf{A}, \mathbf{B}\rangle_{\mathbf{F}}\stackrel{def} {=}\sum_{i, j} \overline{A_{i j}} B_{i j}=\operatorname{Tr}\left(\mathbf{A}^{T} \mathbf{B}\right).
	\end{align}
	After recovering the initial parameters, the optimal rotation optimization can be re-written as,  
	\begin{align}
		&\mathbf{R}^{\star}_{\Base\Lidar} =
		\argmin_{\mathbf{R}_{\Base\Lidar}}\sum_{i=1}^{N}
		\|\mathbf{R}_{\Base\Lidar}\mathbf{R}_{\Base_i} - \mathbf{R}_{\Lidar_i}\|^2_F
		\label{eq:rot_cost_function}\\
		&=
		\resizebox{0.95\columnwidth}{!}{$\underset{\mathbf{R}_{\Base\Lidar}}{\arg
				\min}\sum_{i=1}^{N}\left[\|\mathbf{R}_{\Base_i}^{T}\mathbf{R}_{\Base_i}\| +
			\|\mathbf{R}_{\Lidar_i}^{T}\mathbf{R}_{\Lidar_i}\| -
			2\operatorname{Tr}\left(\mathbf{R}_{\Base\Lidar}\mathbf{R}_{\Base_i}\mathbf{R}_{\Lidar_i}^T\right)\right]$}\nonumber\\
		&= \argmin_{\mathbf{R}_{\Base\Lidar}}\sum_{i=1}^{N}2\left[ 1 -
		\operatorname{Tr}\left(\mathbf{R}_{\Base\Lidar}\mathbf{R}_{\Base_i}\mathbf{R}_{\Lidar_i}^T\right)\right],
		\label{eq:optimal_rotation}
	\end{align}
	where, the rotation components are in $\SEthree$, $\R\R^T = \Identity$.
	Because the constant terms do not affect the $\operatorname{argmin}$ operator, we can discard those and introduce our proposed cost function as, 
	\begin{align}
	J(\mathbf{R}_{\Base\Lidar}) = -\sum_{i=1}^{N}\operatorname{Tr}\left(\mathbf{R}_{\Base\Lidar}\mathbf{R}_{\Base_i}\mathbf{R}_{\Lidar_i}^T\right).
	\label{eq:optimal_rotation_cost_function}
	\end{align}
	\subsubsection{Q-method formulation}
	We formulate our problem as Q-method \cite{davenport1968vector} and follow their ideas to convert the cost function based on rotation matrices to quaternion form as follows,
	\begin{align}
	%&\stackrel{eq. \ref{eq:optimal_rotation_cost_function}} {=}
	%-\sum_{i=1}^{N}\operatorname{Tr}\left(\mathbf{R}_{\Base\Lidar}\mathbf{R}_{\Base_i}\mathbf{R}_{\Lidar_i}^T\right)
	%\nonumber \\
	%&=
	%-\operatorname{Tr}\left[\sum_{i=1}^{N}\left(\mathbf{R}_{\Base\Lidar}\mathbf{R}_{\Base_i}\mathbf{R}_{\Lidar_i}^T\right)\right]\nonumber
	%\\
	\resizebox{0.87\columnwidth}{!}{$J(\mathbf{R}_{\Base\Lidar})
		\stackrel{\eq~\ref{eq:optimal_rotation_cost_function}} {=}
		 -\operatorname{Tr}\left[\mathbf{R}_{\Base\Lidar}
		\sum_{i=1}^{N}\left(\mathbf{R}_{\Base_i}\mathbf{R}_{\Lidar_i}^T\right)\right] = -\operatorname{Tr}\left[\mathbf{R}_{\Base\Lidar}
		\mathbf{\Delta}_{3\times3}\right]
		\label{eq:rotation_cost_function}$},
	\end{align}
	where, $\mathbf{\Delta}_{3\times3} = \sum_{i=1}^{N}\left(\mathbf{R}_{\Base_i}\mathbf{R}_{\Lidar_i}^T\right)$. %often denoted as the attitude profile matrix in
	%aerospace engineering. 
	The rotation matrix can be expressed as a quaternion as,
	\begin{align}
		\mathbf{R}_{\Base\Lidar} =\left(q_{s}^{2}-\mathbf{q}_{v}^{T}
		\mathbf{q}_{v}\right) \mathbf{I}+2 \mathbf{q}_{v} \mathbf{q}_{v}^{T}-2
		q_{s}\left[\mathbf{q}_{v}\right]_\times,
		\label{eq:rot_to_quat}
	\end{align}
	where, $[.]_{\times}$ is the skew-symmetric matrix. % represented as, $
	%[\mathbf{q}_v]_\times=\left[\begin{array}{rrr}0 & -q_z & q_y \\ q_z & 0 & -q_x
	%\\ -q_y & q_x & 0\end{array}\right]$. 
	So, if we rewrite the cost function of \eq~\ref{eq:rotation_cost_function} in quaternion space we get,
	\begin{align}
		&J(\mathbf{q}) 
		\stackrel{\eq~\ref{eq:rot_to_quat}} {=}
		-\operatorname{Tr}\left[\left((q_{s}^{2}-\mathbf{q}_{v}^{T} \mathbf{q}_{v})
		\mathbf{I}+2 \mathbf{q}_{v} \mathbf{q}_{v}^{T}-2
		q_{s}\left[\mathbf{q}_{v}\right]_\times \right)\mathbf{\Delta}\right]
		\label{eq:agrmin_rotation}  \nonumber \\
		%&= \resizebox{0.95\columnwidth}{!}{$-\left(q_{s}^{2}-\mathbf{q}_{v}^{T}
		%	\mathbf{q}_{v}\right) \operatorname{Tr}\left(\mathbf{\Delta}\right)-2
		%	\operatorname{Tr}\left(\mathbf{q}_{v} \mathbf{q}_{v}^{T}
		%	\mathbf{\Delta}\right)+2 q_{s}
		%	\operatorname{Tr}\left[\left(\mathbf{q}_{v}\right)_\times
		%	\mathbf{\Delta}\right]$} \nonumber \\
		%&= \resizebox{0.95\columnwidth}{!}{$-\left(q_{s}^{2}-\mathbf{q}_{v}^{T}
			%\mathbf{q}_{v}\right) \operatorname{Tr}\left(\mathbf{\Delta}\right)-2
			%\left(\mathbf{q}_{v}^{T}\mathbf{\Delta}^T\mathbf{q}_{v} \right)+2 q_{s}
			%\operatorname{Tr}\left[\left(\mathbf{q}_{v}\right)_\times
			%\mathbf{\Delta}\right]$} \nonumber \\
		&= \resizebox{0.95\columnwidth}{!}{$-\left(q_{s}^{2}-\mathbf{q}_{v}^{T}
			\mathbf{q}_{v}\right) \operatorname{Tr}\left(\mathbf{\Delta}\right)-
			\mathbf{q}_{v}^{T}\left(\mathbf{\Delta}^T+\mathbf{\Delta}\right)\mathbf{q}_{v}
			+2 q_{s} \operatorname{Tr}\left[\left(\mathbf{q}_{v}\right)_\times
			\mathbf{\Delta}\right]$}.
	\end{align}
	%\mfallon{the following is too congested. You need to drop something --- especially between eq 21 and eq23}
 Expanding the last term in the \eq~\ref{eq:agrmin_rotation} yields,
	\begin{align}
		&\operatorname{Tr}\left[\left(\mathbf{q}_{v}\right)_\times
		\mathbf{\Delta}\right] \nonumber \\
		&=\resizebox{0.95\columnwidth}{!}{$
			q_x(\mathbf{\Delta}(3,2)-\mathbf{\Delta}(2,3))+q_y(\mathbf{\Delta}(1,3)-\mathbf{\Delta}(3,1))+q_z(\mathbf{\Delta}(2,1)-\mathbf{\Delta}(1,2))$}
		\nonumber \\
		&  = \mathbf{\Lambda}\mathbf{q}_{v},
	\end{align}
	where, $\mathbf{\Lambda}=\left[\begin{array}{ccc} \mathbf{\Delta}(2,3)
		-\mathbf{\Delta}(3, 2) \\ \mathbf{\Delta}(3, 1) - \mathbf{\Delta}(1, 3) \\
		\mathbf{\Delta}(1, 2)-\mathbf{\Delta}(2, 1)\end{array}\right]$. Hence, in quaternion form the cost function can be re-written as,
	\begin{align}
		&\resizebox{0.891\columnwidth}{!}{$J(\mathbf{q}) = -\left(q_{s}^{2}-\mathbf{q}_{v}^{T}
			\mathbf{q}_{v}\right) \operatorname{Tr}\left(\mathbf{\Delta}\right)-
			\mathbf{q}_{v}^{T}\left(\mathbf{\Delta}^T+\mathbf{\Delta}\right)\mathbf{q}_{v}
			+2 q_{s}\mathbf{\Lambda}\mathbf{q}_{v}$}
		\label{eq:quat_cost_fn}
	\end{align}
	Defining, $\mathbf{\Gamma} = \mathbf{\Delta}^T+\mathbf{\Delta}$ and $\mu =
	\operatorname{Tr}(\mathbf{\Delta})$ we get,
	\begin{align}
		&J(\mathbf{q}) \stackrel{\eq~\ref{eq:quat_cost_fn}} {=}
		\resizebox{0.8\columnwidth}{!}{$-\left(q_{s}^{2}-\mathbf{q}_{v}^{T}
			\mathbf{q}_{v}\right) \operatorname{Tr}\left(\mathbf{\Delta}\right)-
			\mathbf{q}_{v}^{T}\left(\mathbf{\Delta}^T+\mathbf{\Delta}\right)\mathbf{q}_{v}
			+2 q_{s}\mathbf{\Lambda}\mathbf{q}_{v}$} \nonumber \\
		&= -\left(\mathbf{q}_{v}^{T}(\mathbf{\Gamma}-\mu \mathbf{I})
		\mathbf{q}_{v}+q_{s} \mathbf{\Lambda} \mathbf{q}_{v}+q_{s} %\mathbf{q}_{v}^{T}
		¨\mathbf{\Lambda}^T+q_{s}^{2} \mu\right) \nonumber \\
		%&= -\left(\mathbf{q}_{v}^{T}(\mathbf{\Gamma}-\mu \mathbf{I}) + q_{s}
		%\mathbf{\Lambda} + \mathbf{q}_{v}^{T} \mathbf{\Lambda}^T+q_{s} \mu\right)
		%\left[\begin{array}{l}\mathbf{q}_{v} \\ q_{s}\end{array}\right] \nonumber\\
		&= -\left[\begin{array}{ll}\mathbf{q}_{v}^{T} &
			q_{s}\end{array}\right]\left[\begin{array}{cc}\mathbf{\Gamma}-\mu \mathbf{I} &
			\mathbf{\Lambda} \\ \mathbf{\Lambda}^{T} & \mu\end{array}\right]
		\left[\begin{array}{l}\mathbf{q}_{v} \\ q_{s}\end{array}\right] \nonumber\\
		&= -\mathbf{q}^T\mathbf{K}\mathbf{q},
		\label{eq:davenport_q_optimization}
	\end{align}
	where, $\mathbf{q}$ is the attitude quaternion (vector first), and $\mathbf{K}$
	is the Davenport matrix denoted as $\mathbf{K} =
	\left[\begin{array}{cc}\mathbf{\Gamma}-\mu \mathbf{I} & \mathbf{\Lambda} \\
		\mathbf{\Lambda}^{T} & \mu\end{array}\right]$. Since, $\mathbf{\Gamma}$ is symmetric, the Davenport matrix is also symmetric. From spectral theorem \cite{hoffman1971linear}, we know that for any symmetric matrix, the eigenvalues are real and the associated eigenvectors form an orthonormal basis. We use this property to ensure a deterministic solution in the next steps of the process.
	
	%\mfallon{if this maths follows the method from Hoffman, then make this clearer - by stating it earlier e.g. `we follow the proposal from'. its a bad idea to imply that you proposed this formulation if you have not.}
	%https://mast.queensu.ca/~br66/419/spectraltheoremproof.pdf
	%https://inst.eecs.berkeley.edu/~ee127/sp21/livebook/thm_sed.html
	
	\subsubsection{Q-method optimization with  Lagrange multiplier}
	We can now perform a constrained optimization by considering
	the orthogonal constraints of the rotation, $\mathbf{q}\mathbf{q}^T =
	\mathbf{I}$, which can be introduced using a Lagrange multiplier, $\lambda$.
	The optimization function for \eq~\ref{eq:davenport_q_optimization} can then
	be re-written as,
	\begin{align}
		\min_{\mathbf{q}, \lambda} J(\mathbf{q}, \lambda) &=-\mathbf{q}^{T} \mathbf{K}
		\mathbf{q}+\lambda\left(\mathbf{q}^{T} \mathbf{q}-1\right)
		\label{eq:davenport_opt}\\
		\text{For minima, }\frac{\partial J}{\partial \mathbf{q}} &= 0 \implies
		\mathbf{K} \mathbf{q} = \lambda\mathbf{q}.
		\label{eq:davenport_lagrange}
	\end{align}
	The attitude quaternion which minimizes the cost function is the unit
	eigenvector corresponding to the largest eigenvalue of $\mathbf{K}$ (eigenvalues of a symmetric matrix are real and hence can be sorted).
	\begin{align}
		\mathbf{q}^{\star} &=
		\max_{\operatorname{Eigenvalues}(\mathbf{K})}{\operatorname{Eigenvectors}(\mathbf{K})}
		\label{eq:R_estimated}\\
		\implies \min_{\mathbf{q}, \lambda} J(\mathbf{q}, \lambda) 
		%&= -\mathbf{q}^{T}
		%\mathbf{K} \mathbf{q}+\lambda\left(\mathbf{q}^{T} \mathbf{q}-1\right) \nonumber
		%\\
		&\stackrel{\eq~\ref{eq:davenport_opt}} {=} -\lambda\mathbf{I} + \lambda(\mathbf{I} - 1) = -\lambda.
	\end{align}
	We only update the estimated extrinsic rotation parameters if it improves the rotational cost function, as described in \ref{sec:cost_function}.  
	
	\subsubsection{Translation estimation}
	Note that, because the motion of the lidar is estimated in the
	sensor frame, it is not possible to solve for the translation component without the relative transformations and the rigid body constraint provided by Hand-Eye method \cite{hand-eye_tsai_1989}.
	\begin{figure}[!h]
		\centering
		\vspace{-2mm}
		\includegraphics[width=1\columnwidth]{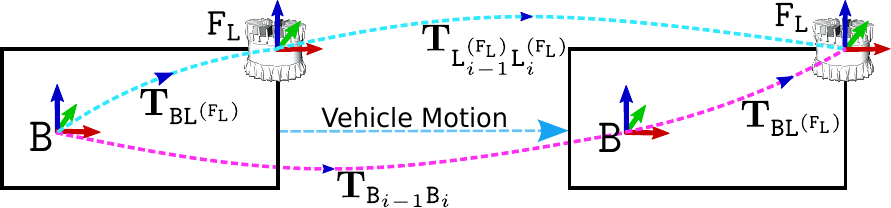}
		\vspace{-7mm}
		\caption{Illustration of Hand-eye calibration between $\Base$ and $\mathtt{L}^{(\mathtt{F}_\mathtt{L})}$ frame.} 
		\label{fig:hand-eye}
	\end{figure}
		From \Figure \ref{fig:hand-eye} we can write:
		\begin{equation}
		{\T}{_{{\Base_{i-1}\Base_i}}}{\T}{_{\Base\Lidar}} = {\T}{_{\Base\Lidar}}{\T}{_{\Lidar_{i-1}\Lidar_i}},
		\label{eq:hand-eye}
		\end{equation} 
		
%		\begin{equation}
%			\begin{aligned}
%			{\T}{_{\mathtt{F}{\mathtt{R}}}}{\T}{_{{\mathtt{R}_{i-1}\mathtt{R}_i}}} &= {\T}{_{{\mathtt{F}_{i-1}\mathtt{F}_i}}}{\T}{_{\mathtt{F}{\mathtt{R}_i}}}\\
%			{\T}{_{\mathtt{F}{\mathtt{R}}}}{\T}{_{{\mathtt{R}_{i-1}\mathtt{R}_i}}} & = {\T}{_{{\mathtt{F}_{i-1}\mathtt{F}_i}}}{\T}{_{\mathtt{F}{\mathtt{R}}}}{\T}{_{\mathtt{R}^\prime_i{\mathtt{R}_i}}}
%			\end{aligned}
%		\end{equation} 
		
		\noindent which can be further decomposed into the translation part with the
	pre-computed optimal rotation in previous step as,
		\begin{equation}
			\underbrace{\left(\mathbf{R}_{{\Base_{i-1}}{\Base_i}} - \mathbf{I}\right)}_{\mathbf{A}_i}\mathbf{t}{_{\Base\Lidar}} \stackrel{\eq~\ref{eq:R_estimated}} {=} \underbrace{\left(\mathbf{R}{_{\Base\Lidar}^{\star}}\mathbf{t}_{{\Lidar_{i-1}}{\Lidar_i}} -
	\mathbf{t}_{{\Base_{i-1}}{\Base_i}} \right)}_{\mathbf{B}_i},
			\label{eq:hand-eye-t}
		\end{equation}
	where, $\forall i=1(1)N$, which can be computed using similar principle of solving \Equation\ref{eq:bvls} with known prior bounds.

	\begin{algorithm}[!bht]
		\DontPrintSemicolon
		\KwInput{Lidar point cloud in $\Lidar$ frame, GNSS reference poses in $\Base$ frame = $\T_{\Base_i}$, Angular velocity signals of IMUs embedded in lidar and GNSS = $\{\boldsymbol{\omega}_{\Imu_i}, \boldsymbol{\omega}_{\Base_i}\}$, Batch size = $N$, Pose error threshold = $\beta$, Information threshold = $\epsilon$.}

		\KwOutput{$\T^\star_{\Base\Lidar}$.}
		
		\If{not initialized}
		{
			
			\If{$\T_{\Lidar\Imu}$ is unknown}
			{
				\tcp{Estimate $\T_{\Lidar_1\Lidar_N}$ with lidar odometry and $\T_{\Imu_1\Imu_N}$ using \eq~\ref{eq:imu-preintegration} for Hand-Eye}
				
				Estimate $\mathbf{T}_{\Lidar\Imu}$ using \eq~\ref{eq:rot_lidar_imu} and \eq~\ref{eq:hand-eye-t-li}.
			}
		
			Estimate $\T_{\Lidar_i}$ using lidar odometry.
			
			\tcp{Since, initial $\mathbf{t}_{\Base\Lidar}$ can be measured, only initial $\mathbf{R}_{\Base\Lidar}$ 
				computation is necessary}
			
			Compute initial $\mathbf{R}_{\Base\Lidar}$ and $\mathbf{t}_{\Base\Lidar}$ by aligning the set of $\tran_{\Lidar_i}$ \textit{wrt} $\tran_{\Base_i}$, $\forall i=1(1)N$ by Umeyama \cite{umeyama1991least} or Horn \cite{horn1987closed} alignment.
		}
		
		Let, $\mathtt{Error}({\T}_{\Base\Lidar}) \stackrel{\eq~\ref{eq:error_term}} {=} \frac{1}{N}\sqrt{\sum_{i=1}^{N} \xi_{\T_{\Base\Lidar}}({\T}{_{\Base_i}},{\T}{_{\Lidar_i}})}$
		
		and, $\mathtt{R_{Error}}({\R}_{\Base\Lidar}) \stackrel{\eq~\ref{eq:error_term}} {=} \frac{1}{N}\sqrt{\sum_{i=1}^{N} \xi_{\R_{\Base\Lidar}}({\R}{_{\Base_i}},{\R}{_{\Lidar_i}})}$.
		
		Let, ${\T}_{\Base} = \{\}$ and ${\T}_{\Lidar} = \{\}$, be the set of all GNSS and lidar poses which satisfy observability criteria.  
		
		\While{$\mathtt{Error}(\T_{\Base\Lidar}) \ge \beta$}
		{
			Estimate $\T_{\Lidar_i}$ using lidar odometry.

			Find closest $\T_{\Base_i}$ \textit{wrt} $\T_{\Lidar_i}$ based on timestamp correspondance of the poses.
			
			\If{No. of corresponding poses = $N$}
			{	
				
				Compute $\mathcal{I}_{N\times N}(\boldsymbol{\omega}_{\Base_i}, \boldsymbol{\omega}_{\Imu_i})$ using \eq~\ref{eq:residual_calc}.
				
				\If{Minimum singular value of $\mathcal{I}_{N\times N}$ $\geq$ $\epsilon$}
				{
					
					$\T_{\Base} := \T_{\Base} \cup  \{\T_{\Base_i}\}$, $\forall i=1(1)N$.
					
					$\T_{\Lidar} := \T_{\Lidar} \cup  \{\T_{\Lidar_i}\}$, $\forall i=1(1)N$.
					
					Compute $\hat{\mathbf{R}}_{\Base\Lidar}$ between $\T_{\Base}$ and $\T_{\Lidar}$ by solving for $\mathbf{K}$ using \eq~\ref{eq:R_estimated}.% and retain it for the next if it improves rotational cost of \eq~\ref{eq:error_term}. %by using $\mathbf{R}_{\Base\Lidar}$ and $\hat{\mathbf{t}}_{\Base\Lidar^{(k)}}$.
					
					\If{$\mathtt{R_{Error}}(\hat{\R}_{\Base\Lidar}) \le \mathtt{R_{Error}}({\R}_{\Base\Lidar})$}
					{
						$\R_{\Base\Lidar} := \hat{\R}_{\Base\Lidar}$.
						
						Solve for $\hat{\mathbf{t}}$ using \eq~\ref{eq:hand-eye-t}.
						
						%\resizebox{0.73\columnwidth}{!}{$\mathtt{Error}(\hat{\T}_{\Base\Lidar}) \stackrel{\eq~\ref{eq:error_term}} {=} \frac{1}{N}\sqrt{\sum_{i=1}^{N} \xi_{\hat{\T}_{\Base\Lidar}}({\T}{_{\Base_i}},{\T}{_{\Lidar_i}})}$}
						
						%\tcp{If $\mathtt{RMSE}_i$ does not improve, re-use the current poses in the next batch and discard currently estimated extrinsics}
						\If{$\mathtt{Error}(\hat{\T}_{\Base\Lidar})$  $\le \mathtt{Error}(\T_{\Base\Lidar})$}
						{
							\tcp{Update current extrinsics}
							
							$\T_{\Base\Lidar} := \hat{\T}_{\Base\Lidar}$.
						}	
					}
				}
				\Else
				{
					Discard the set of current $N$ poses.
				}
			}
		}
		$\T^\star_{\Base\Lidar} := {\T}_{\Base\Lidar}$
		
		\caption{Online lidar extrinsic calibration}
		\label{algo:online-extrinsics}
	\end{algorithm}
	\setlength{\textfloatsep}{7pt}% Remove \textfloatsep

	\subsubsection{Observability analysis}
	Not all motions of the test vehicle sufficiently excite the degrees of freedom of the calibration technique. To ensure that online calibration is reliable it is necessary to identify the dataset segments which is suitable and ignore the others.
	
	The extrinsics of $\Imu^{(k)}$ \textit{wrt} $\Base$ frame is strongly correlated with the extrinsics of $\Lidar^{(k)}$ \textit{wrt} $\Base$ frame. Moreover, the angular velocity at all the points in a rigid body are identical and can be correlated to the motion estimate of the vehicle. Thus, we formulate a cost function for maximizing the likelihood of matching the angular velocity between $\Imu^{(k)}$ and $\Base$ frame as,
	\begin{equation}
		\begin{aligned}
			\R_{\Base\Imu}^{\star} &= \argmin_{\R_{\Base\imu}} \sum_{i=1}^{N}
			\| \R_{\Base\imu} \boldsymbol{\omega}_{\Base_i} - \boldsymbol{\omega}_{\Imu_i} \|^2_{\boldsymbol{\sum_i}}, \\
			& \textit{s.t.} \ \ \R_{\Base\Imu} \R_{\Base\Imu}^T = \Identity_3,
			\label{eq:imu_R_calibration}
		\end{aligned}
	\end{equation}
	where, $\boldsymbol{\omega}_{\Imu_i^{(k)}}$ and $\boldsymbol{\omega}_{\Base_i}$ are the angular velocities of $\Imu^{(k)}$ and $\Base$ at timestamp $t_i$. Since,  \eq~\ref{eq:imu_R_calibration} has similar form as \eq~\ref{eq:unbounded} we solve it iteratively with the update step as,
	\begin{equation}
		\begin{aligned}
			\underbrace{\left(\sum_{i} \mathbf{J}_{i}^{\top} \Sigma_{i}^{-1} \mathbf{J}_{i}\right)}_{\text{Fisher information matrix}} \mathbf{\Delta}{\mathbf{\R}}_{\Base\imu}& \stackrel{\eq~\ref{eq:bvls_iterative_solution}} {=} -\sum_{i} \mathbf{J}_{i}^{\top} \Sigma_{i}^{-1} (\hat{\mathbf{\R}}_{\Base\imu} \boldsymbol{\omega}_{\Base_i} - \boldsymbol{\omega}_{\Imu_i})
			\label{eq:residual_calc}
		\end{aligned}
	\end{equation}
	where, ${\boldsymbol{\Sigma}_i} = \operatorname{cov}(\hat{\mathbf{\R}}_{\Base\imu} \boldsymbol{\omega}_{\Base_i} - \boldsymbol{\omega}_{\Imu_i})$ and $\mathbf{J}_{i} = \boldsymbol{\omega}_{\Base_i}^T$. The Fisher information matrix, $\mathcal{I}_{N\times N}$ captures all the information contained in the measurements. Since, we compute this online we do processing in a pre-defined batch size of $N$. We perform a Singular Value Decomposition of $\mathcal{I}_{N\times N}$ for each batch as:
	\begin{equation}
		\mathcal{I}_{N\times N} = \textbf{USU}^T,
	\end{equation}
	where, $\textbf{U} = [\textbf{u}_1, \textbf{u}_2 \dots , \textbf{u}_N]$ and $\textbf{S} = \text{diag}(\sigma_1, \sigma_2 \dots , \sigma_N)$ is a diagonal matrix of singular values in decreasing order. The value of the minimum singular value gives an indication about the information in the batch data. If the minimum singular value is more than an threshold (design choice) we conclude the data in the batch is sufficiently excited for reliable extrinsics computation. If the batch of IMU signals provide enough information, the estimated poses from the lidar odometry are selected for online calibration computation as described in Algorithm \ref{algo:online-extrinsics}. 

	\subsubsection{Stopping criteria}
	\label{sec:cost_function}
	In our Algorithm \ref{algo:online-extrinsics}, we  trigger the optimization process when a pre-defined batch of poses have been accumulated. However, instead of considering only the current batch of poses we consider the full set of accumulated poses from the beginning of operation to encapsulate wider excitation of different degrees of freedom. At each iteration we compute the normalized cost function and update the estimated extrinsic parameters, if it improves the cost function. Based on our empirical analysis we stop the process once our cost function is below a certain threshold. We define our cost function in \SEthree using properties from Riemannian geometry \cite{petersen2006riemannian} as,
	\begin{align}
		&\resizebox{0.88\columnwidth}{!}{$\xi_{{\T}^\star_{\Base\Lidar}}({\T}{_{\Base_i}},{\T}{_{\Lidar_i}})	= \underbrace{\|\log(\mathbf{R}{_{\Base\Lidar}^{\star}}\mathbf{R}{_{\Base_i}}\mathbf{R}{_{\Lidar_i}^T})\|^2}_{\mathtt{Rotation\ Cost} = \xi_{{\R}_{\Base\Lidar}}} + \underbrace{\|\mathbf{R}{_{\Base\Lidar}^{\star}}\tran_{\Base_i} + \tran_{\Base\Lidar}^\star - \tran_{\Lidar_i}\|^2}_{\mathtt{Translation \ Cost} = \xi_{{\tran}_{\Base\Lidar}}}$}, \label{eq:error_term}\\
		&\text{where, } \log (\R)=\left\{\begin{array}{cc}0, & \theta=0 \\ \frac{\theta}{2 \sin \theta}\left(\R-\R^{\top}\right) & |\theta| \in(0, \pi)\end{array}\right. \nonumber \\
		&\text{and, } \operatorname{Tr}(\R) = 1 + 2 \cos\theta. \nonumber
	\end{align}

	\subsubsection{Sensitivity analysis}
	Here we provide an analytical study of the effect of additive noise on our optimization problem.
	
	(i) \textit{Rotation noise component}: First we consider the rotational noisy components. We assume that the additive noise still constructs a rotation matrix in \SEthree. Considering the GNSS and the estimated lidar rotational components with noise as $\R_{\Base_i} + \boldsymbol{\eta}_{\Base_i}$ and $\R_{\Base\Lidar_i} +\boldsymbol{\eta}_{\Lidar_i}$, our optimization problem becomes,
	\begin{align}
		\R_{\Base\Lidar}^{\star} &= \argmin_{\R_{\Base\Lidar}, \boldsymbol{\eta}_{\Base_i}, \boldsymbol{\eta}_{\Lidar_i} }\sum_{i=1}^{N}
		\left(\|\boldsymbol{\eta}_{\Base_i}\|^2_{\Sigma^{-1}_{\Base\Base}} +  \|\boldsymbol{\eta}_{\Lidar_i}\|^2_{\Sigma^{-1}_{\Lidar\Lidar}} \right) \\
		& \textit{s.t.} \ \ \R_{\Base\Lidar}(\R_{\Base_i} + \boldsymbol{\eta}_{\Base_i} ) =
		(\R_{\Lidar_i} + \boldsymbol{\eta}_{\Lidar_i}) \;\; \forall i.
		\label{eq:noise_rot_eq}
	\end{align}
	Thus, rearranging \eq~\ref{eq:noise_rot_eq} would yield, 
	\begin{equation}
		\boldsymbol{\eta}_{\Lidar_i} =
		(\R_{\Base\Lidar}\R_{\Base_i} - \R_{\Lidar_i}) + \R_{\Base\Lidar}\boldsymbol{\eta}_{\Base_i}.
	\end{equation}
	Considering equal co-variance norms we can re-write the error term, $\Xi$ at $t_i$ as,
	\begin{align}
		&\Xi(\boldsymbol{\eta}_{\Base_i}, \boldsymbol{\eta}_{\Lidar_i}) = \|\boldsymbol{\eta}_{\Base_i}\|^2 +  \|\boldsymbol{\eta}_{\Lidar_i}\|^2 \\
		%&= \|\boldsymbol{\eta}_{\Base_i}\|^2 +  \| (\R_{\Base\Lidar}\R_{\Base_i} - \R_{\Lidar_i}) + \R_{\Base\Lidar}\boldsymbol{\eta}_{\Base_i}\|^2 + \|\R_{\Base\Lidar}\boldsymbol{\eta}_{\Base_i}\|^2 \nonumber \\
		&= \resizebox{0.89\columnwidth}{!}{$ 2\|\boldsymbol{\eta}_{\Base_i}\|^2 +  2 (\R_{\Base\Lidar}\R_{\Base_i} - \R_{\Lidar_i})^T\R_{\Base\Lidar}\boldsymbol{\eta}_{\Base_i} + \|\R_{\Base\Lidar}\R_{\Base_i} - \R_{\Lidar_i}\|^2$}, \nonumber \\
		&\resizebox{0.99\columnwidth}{!}{$\text{since}, \|\R_{\Base\Lidar}\boldsymbol{\eta}_{\Base_i}\|^2 = (\R_{\Base\Lidar}\boldsymbol{\eta}_{\Base_i})^T\R_{\Base\Lidar}\boldsymbol{\eta}_{\Base_i}  =\boldsymbol{\eta}_{\Base_i}^T\R_{\Base\Lidar}^T\R_{\Base\Lidar}\boldsymbol{\eta}_{\Base_i} = \|\boldsymbol{\eta}_{\Base_i}\|^2$} \nonumber.
	\end{align}
	To minimize, we equate the Jacobian of $\Xi(\boldsymbol{\eta}_{\Base_i}, \boldsymbol{\eta}_{\Lidar_i})$ to zero.
	\begin{align}
		&\implies &\frac{\partial \Xi}{\partial \boldsymbol{\eta}_{\Base_i}} = 0 \nonumber \\
		&\text{or, } &4\boldsymbol{\eta}_{\Base_i} + 2\R_{\Base\Lidar}^T(\R_{\Base\Lidar}\R_{\Base_i} - \R_{\Lidar_i}) = 0 \nonumber \\
		&\text{or, } &\boldsymbol{\eta}_{\Base_i} = -\frac{1}{2}(\R_{\Base_i} - \R_{\Base\Lidar}^{-1}\R_{\Lidar_i}).
	\end{align}
	So, it turns out that under noisy conditions the optimal rotation finding problem has a similar form as our original rotation optimization problem in \eq~\ref{eq:rot_cost_function}.
	
	(ii) \textit{Translation noise component}:
	Let the relative noisy translation components for the lidar and the GNSS measurements between timestamps $t_{i-1}$ and $t_i$ be $\boldsymbol{\delta}_{\Lidar_i}$ and $\boldsymbol{\delta}_{\Base_i}$. Let, $\boldsymbol{\tau}_{\Lidar_i} = \mathbf{t}_{{\Lidar_{i-1}}{\Lidar_i}}$ and $\boldsymbol{\tau}_{\Base_i} = \mathbf{t}_{{\Base_{i-1}}{\Base_i}}$. Our optimization problem becomes,
	
	\begin{align}
		&\tran_{\Base\Lidar}^{\star} = \argmin_{\tran_{\Base\Lidar}, \boldsymbol{\delta}_{\Base_i}, \boldsymbol{\delta}_{\Lidar_i} }\sum_{i=1}^{N-1}
		\left(\|\boldsymbol{\delta}_{\Base_i}\|^2_{\Sigma^{-1}_{\Base\Base}} +  \|\boldsymbol{\delta}_{\Lidar_i}\|^2_{\Sigma^{-1}_{\Lidar\Lidar}} \right) \\
		& \resizebox{0.89\columnwidth}{!}{$\textit{s.t.} \ \ (\mathbf{R}_{{\Base_{i-1}}{\Base_i}} - \mathbf{I})\mathbf{t}{_{\Base\Lidar}} 
			 = \mathbf{R}{_{\Base\Lidar}}(\tau_{\Lidar_i} + \boldsymbol{\delta}_{\Lidar_i}) +
			(\tau_{\Base_i} + \boldsymbol{\delta}_{\Base_i})  \;\; \forall i$.}
		\label{eq:noise_tran_eq}
	\end{align}
	Rearranging \eq~\ref{eq:noise_tran_eq} we get,
	\begin{align}
		 \resizebox{0.89\columnwidth}{!}{$\boldsymbol{\delta}_{\Lidar_i} = \mathbf{R}{_{\Base\Lidar}^T}(\mathbf{R}_{{\Base_{i-1}}{\Base_i}} - \mathbf{I})\mathbf{t}{_{\Base\Lidar}} + \mathbf{R}{_{\Base\Lidar}^T}(\tau_{\Base_i} + \boldsymbol{\delta}_{\Base_i}) - \tau_{\Lidar_i}.$}
	\end{align}
	Considering equal co-variance norms we can minimize the error term, $\Xi(\boldsymbol{\delta}_{\Base_i}, \boldsymbol{\delta}_{\Lidar_i})$ by setting the Jacobians equal to zero.
	\begin{align}
		&\implies \frac{\partial \Xi}{\partial \boldsymbol{\delta}_{\Base_i}} = 0 \nonumber \\
		&\text{or, } \resizebox{0.89\columnwidth}{!}{$2\boldsymbol{\delta}_{\Base_i} + 2\R_{\Base\Lidar}\left[\mathbf{R}{_{\Base\Lidar}^T}(\mathbf{R}_{{\Base_{i-1}}{\Base_i}} - \mathbf{I})\mathbf{t}{_{\Base\Lidar}} + \mathbf{R}{_{\Base\Lidar}^T}(\tau_{\Base_i} + \boldsymbol{\delta}_{\Base_i}) - \tau_{\Lidar_i}\right] = 0$} \nonumber \\
		&\text{or, } 2\boldsymbol{\delta}_{\Base_i} + (\mathbf{R}_{{\Base_{i-1}}{\Base_i}} - \mathbf{I})\mathbf{t}{_{\Base\Lidar}} - (\mathbf{R}{_{\Base\Lidar}}\tau_{\Lidar_i} - \tau_{\Base_i}) = 0 \nonumber \\
		&\text{or, } \boldsymbol{\delta}_{\Base_i} = -\frac{1}{2}\left[(\mathbf{R}_{{\Base_{i-1}}{\Base_i}} - \mathbf{I})\mathbf{t}{_{\Base\Lidar}} - (\mathbf{R}{_{\Base\Lidar}}\mathbf{t}_{{\Lidar_{i-1}}{\Lidar_i}} - \mathbf{t}_{{\Base_{i-1}}{\Base_i}}) \right].
	\end{align}
	Thus, for the translation case we can also show that the original optimization problem described in \eq~\ref{eq:hand-eye-t} holds true under noisy conditions as well.

	\section{Experimental Results}
	To demonstrate the perforance of the online calibration system, we conducted several experiments using real world data collected from the two vehicles illustrated in \Figure\ref{fig:coordinate-frames} which have a different configuration.

	%\mfallon{use `illustrated in Fig XXX'}
	
	\subsection{Dataset}
	\label{sec:dataset}
	\begin{table}[!h]
		\centering
		%\vspace{2mm}
		%\fontsize{18}{18}\selectfont
		\resizebox{\columnwidth}{!}{
			\begin{tabular}{l|cccc}  \toprule
				%\begin{tabular*}{\textwidth}{@{\extracolsep{\fill}}l|cccc}\toprule
				\multicolumn{5}{c}{Dataset collection details for the experiments}\\
				\midrule \midrule
				\textbf{Data} & Vehicle & Sensor setup & Length (\si{Km}) & Duration (\si{secs})\\
				\midrule
				Seq-1 & Bus & $\mathtt{F}_\mathtt{L}, \mathtt{F}_\mathtt{R}, \mathtt{R}_\mathtt{L}, \mathtt{R}_\mathtt{R}$ & 1.473 & 136 \\
				Seq-2 & Bus & $\mathtt{F}_\mathtt{L}, \mathtt{F}_\mathtt{R}$ & 2.086 & 135 \\
				Seq-3 & Bus & $\mathtt{F}_\mathtt{L}, \mathtt{F}_\mathtt{R}$ & 1.089 & 50 \\
				Seq-4 & Truck & $\mathtt{F}_\mathtt{L}, \mathtt{F}_\mathtt{R}, \mathtt{R}_\mathtt{L}, \mathtt{R}_\mathtt{R}$ & 0.478 & 67 \\
				Seq-5 & Truck & $\mathtt{F}_\mathtt{L}, \mathtt{F}_\mathtt{R}, \mathtt{R}_\mathtt{L}, \mathtt{R}_\mathtt{R}$ & 1.212 & 73 \\
				Seq-6\textsuperscript{\dag} & Truck & $\mathtt{F}_\mathtt{L}-\text{Top}, \mathtt{F}_\mathtt{R}-\text{Top}$ & 0.224 & 164 \\
				Seq-7\textsuperscript{\dag} & Truck & $\mathtt{F}_\mathtt{L}-\text{Top}, \mathtt{F}_\mathtt{R}-\text{Top}$ & 1.088 & 263 \\
				\bottomrule
				\multicolumn{5}{l}{\textsuperscript{\dag}\textit{Front 2 top lidars mounted on the cab were used here.}}
		\end{tabular}}
		\vspace{-3mm}
		\caption{}
		\label{tab:dataset}
		\vspace{-4mm}
	\end{table}
	We use data from a GNSS receiver to produce ground truth (GT) motion trajectories of the $\Base$ frame. We analyze our results on 7 collected test sequences described in Table~\ref{tab:dataset}. 

	\subsection{Online calibration results and repeatability analysis}
	For the online calibration process we start from a set of un-calibrated lidars for which only the translation components from the vehicle CAD parameters are known. A GT calibration is performed offline by refining the vehicle CAD parameters. The refinement process analyzes known static feature positions around the vehicle and matches the corresponding detected features between the lidars. We compare our results to these GT parameters for both vehicles. We measured translation and rotation error as:   
	\begin{align}
		\Delta t &= \frac{1}{3} \sqrt{\| \widehat{\mathbf{t}} - \mathbf{t}_{\operatorname{GT}}\|^2_F} , \\
		\Delta R &=
		\frac{180}{\pi}\operatorname{cos}^{-1}\left[\frac{1}{2}\left(\operatorname{Tr}(\widehat{\mathbf{R}}^{-1}\mathbf{R}_{\operatorname{GT}})-1\right)\right],
	\end{align}
	where, $\Delta R$ is the rotation along the principle eigenvector of $(\widehat{\mathbf{R}}^{-1}\mathbf{R}_{\operatorname{GT}})$. We summarize our results in Table \ref{tab:calib_results} and compare our performance to the Kabsch alignment \cite{Kabsch1978} technique. 

	In Table \ref{tab:calib_results}, we show the average of $\Delta t$ and $\Delta R$ for the bus and truck sequences. It can be seen that our online method produces better results for translation components because of the bounded variable least square optimization \ref{eq:bvls}. Our method also produces comparable results for rotation components outperforming Kabsh alignment in most scenarios. 
	\normalsize
	\begin{table}[!h]
		\centering
		
		\resizebox{0.9\columnwidth}{!}{
			\begin{tabular}{l|c|cccc}  \toprule
				\multicolumn{6}{c}{Average performance comparison of calibration routines}\\
				\midrule \midrule
				\multirow{2}{*}{\raisebox{-\heavyrulewidth}{\textbf{Vehicle}}} &
				\multirow{2}{*}{\raisebox{-\heavyrulewidth}{Lidar position}} &
				\multicolumn{2}{c}{Kabsch alignment} & \multicolumn{2}{c}{Ours online method}
				\\
				\cmidrule(lr){3-6}
				& & $\Delta t$ [\si{\metre}] & $\Delta R$ [\si{\degree}] & $\Delta t$ [\si{\metre}] & $\Delta R$ [\si{\degree}] \\
				\midrule
				\multirow{4}{*}{\raisebox{-\heavyrulewidth}{Bus}} 
				& $\mathtt{F}_\mathtt{L}$ & 10.603 & \textbf{4.370} & \textbf{0.766} & 4.619\\
				& $\mathtt{F}_\mathtt{R}$ &  16.484 & 2.167 & \textbf{0.333} & \textbf{2.069}\\
				& $\mathtt{R}_\mathtt{R}$ & 3.301 & \textbf{0.249} & \textbf{0.413} & 0.285\\
				& $\mathtt{R}_\mathtt{L}$ & 3.246 & 0.257 & \textbf{0.359} & \textbf{0.500}\\
				\midrule
				\multirow{4}{*}{\raisebox{-\heavyrulewidth}{Truck}} 
				& $\mathtt{F}_\mathtt{L}$ & 9.384 & 11.968 & \textbf{0.318} & \textbf{5.301}\\
				& $\mathtt{F}_\mathtt{R}$ & 8.583 & 2.566 & \textbf{0.340} & \textbf{2.006}\\
				& $\mathtt{R}_\mathtt{R}$ & 2.536 & \textbf{1.192} & \textbf{0.311} & 2.113\\
				& $\mathtt{R}_\mathtt{L}$ & 2.453 & \textbf{0.594} & \textbf{0.506} & 1.118\\
				& $\mathtt{F}_\mathtt{L}-\mathtt{Top}$ & 9.637 & 2.462 & \textbf{0.346} & \textbf{2.091}\\
				& $\mathtt{F}_\mathtt{R}-\mathtt{Top}$ & 8.907 & 3.387 & \textbf{0.388} & \textbf{2.141}\\
				\bottomrule
		\end{tabular}}
		\caption{}
		\label{tab:calib_results}
		\vspace{-8mm}
	\end{table}
	
	\begin{figure}[!h]
		\centering
		\vspace{-2mm}
		\includegraphics[width=1\columnwidth]{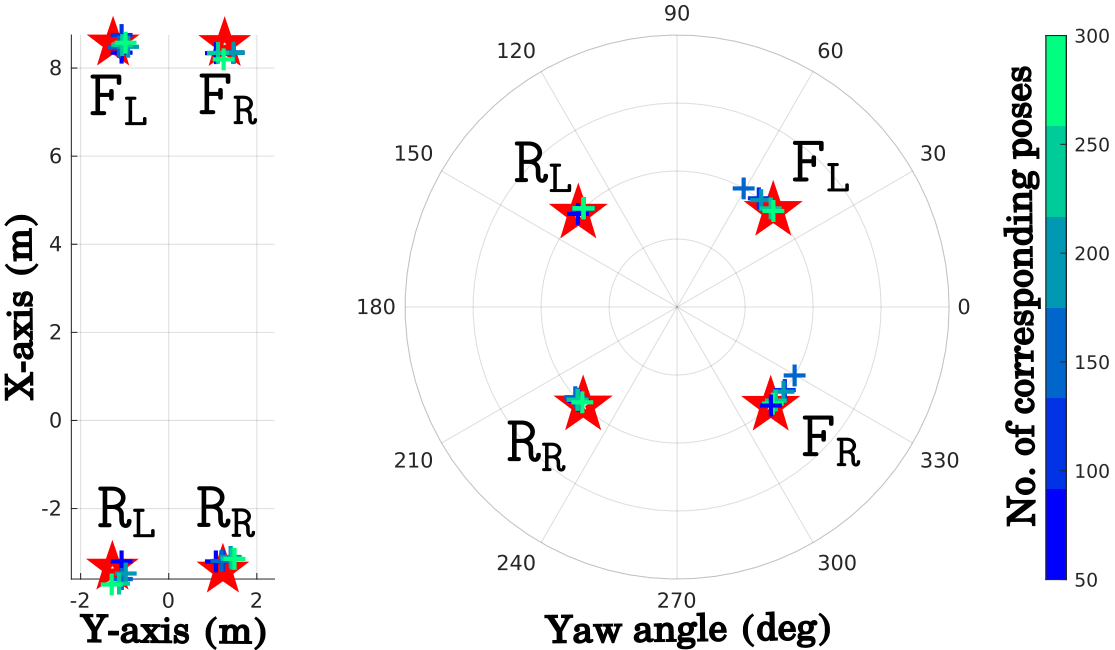}
		\vspace{-6mm}
		\caption{Estimated $t_x$, $t_y$ and $\texttt{Yaw}$ for bus sequence-1 over different pose sizes. The GT calibration parameters are represented in red color.}
		\label{fig:calibration-bus}
	\end{figure}
	\begin{figure}[!h]
		\centering
		\includegraphics[width=1\columnwidth]{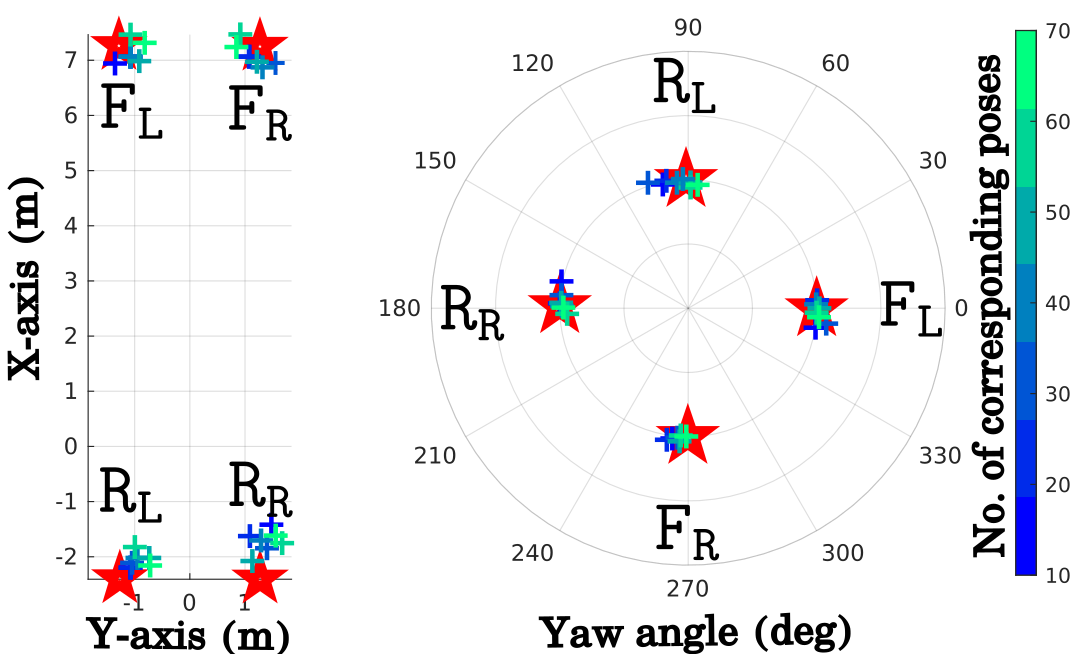}
		\vspace{-6mm}
		\caption{Estimated $t_x$, $t_y$ and $\texttt{Yaw}$ for truck sequence-4 over different pose sizes. The GT calibration parameters are represented in red color.}
		\label{fig:calibration-truck}
	\end{figure}
	We also show the calibration results for the individual transformation components in \Figure\ref{fig:calibration-bus} for bus and \Figure\ref{fig:calibration-truck} truck for different trajectories. Since, we compare the estimated lidar pose in sensor frame to the corresponding GNSS pose in $\Base$ frame for calibration we study the effect of the size of the associated corresponding poses used for calibration. 

	As seen in both the sequences, the rotation components converges to the prior GT calibration as the number of corresponding poses in the optimization increases. We solved for the translation components using a bounded variable problem ($\pm 0.3\si{m}$) since the initial CAD estimates are reliable and found it converged towards the GT parameters as well.
	
\begin{figure}[!h]
		\vspace{-3mm}
		\centering
		\includegraphics[width=1\columnwidth]{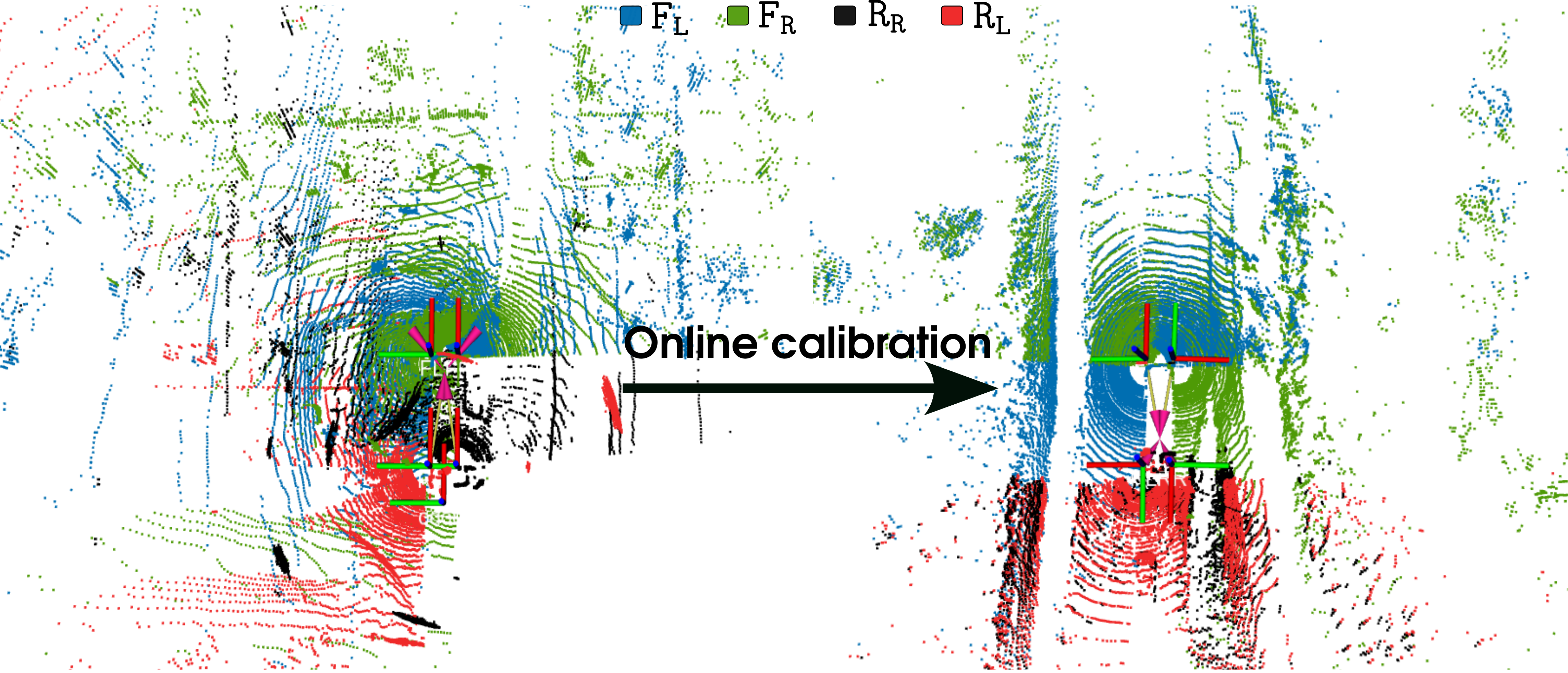}
		\vspace{-5mm}
		\caption{Online calibration results for sequence-4.}
		\vspace{-3mm}
		\label{fig:calibration-result}
	\end{figure}

	After running our algorithm we calibrate the lidars online as seen in \Figure\ref{fig:calibration-result}.	Additionally, we demonstrate the online calibration process in real-time in our supplementary video.

	\subsection{Observability analysis}
	
	For our observability analysis we extract the information matrix from the angular velocity signals. 
	\begin{figure}[!h]
		\vspace{-1mm}
		\centering
		\includegraphics[width=1\columnwidth]{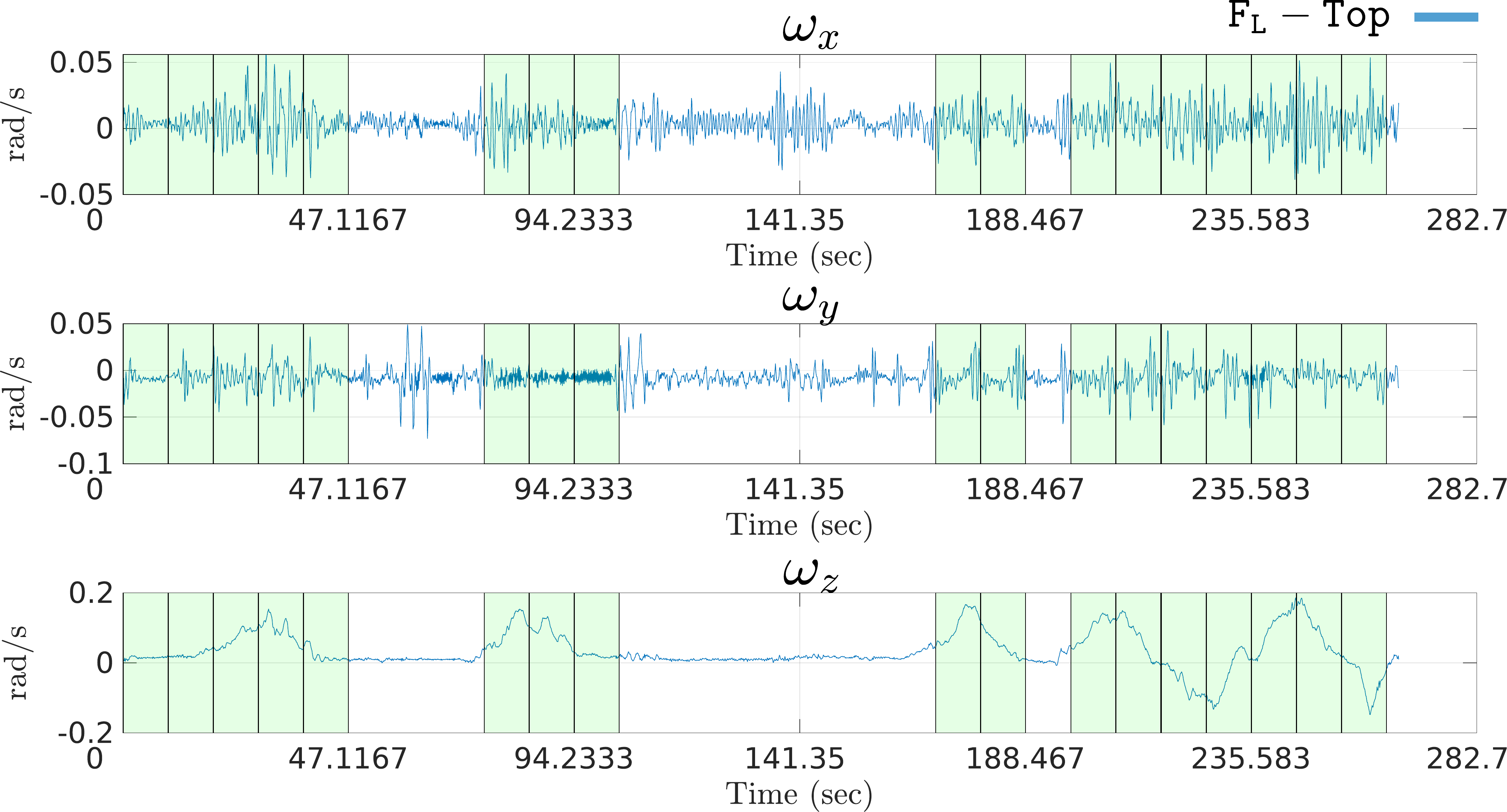}
		\vspace{-7mm}
		\caption{Illustration of observability analysis for Seq-7 with angular velocity.}
		\label{fig:obs_ang_vel}
	\end{figure}
	We only choose to estimate calibration in segments for which there is enough excitation in the angular velocity. These segments are highlighted in green
    in \Figure\ref{fig:obs_ang_vel}.

\begin{figure}[!h]
		\centering
		\vspace{-2mm}
		\includegraphics[width=1\columnwidth]{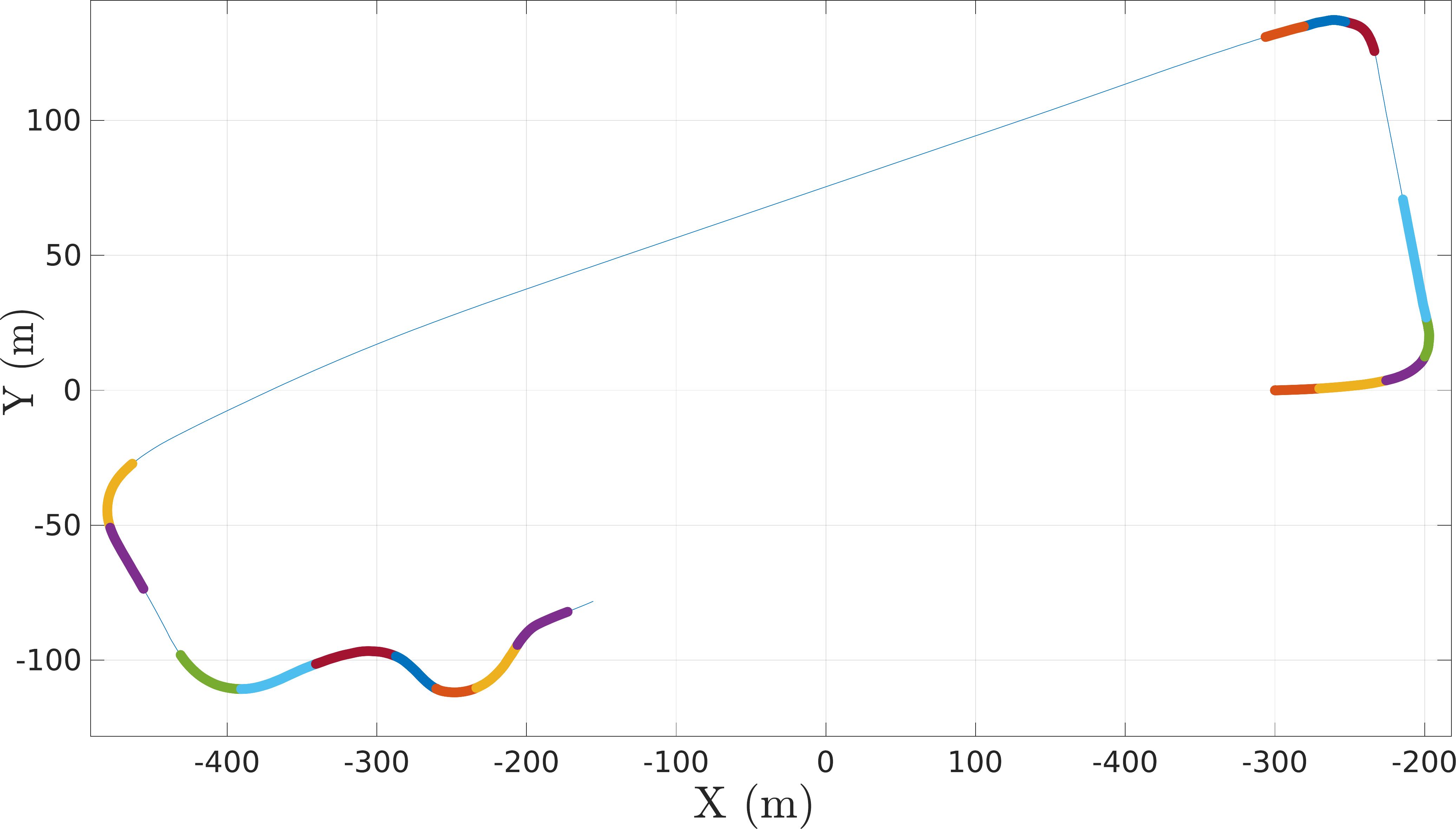}
		\vspace{-6mm}
		\caption{Illustration of selected poses with observability analysis for Seq-7.}
		\label{fig:obs_traj}
		\vspace{-5mm}
	\end{figure}

	The corresponding trajectory is shown in \Figure\ref{fig:obs_traj}. We can see that the trajectory segments selected for calibration correspond are those with significant turning maneuvers.
	
	\section{Conclusion}
	\label{sec:conclusion}
	We have presented an observability-aware online multi-lidar calibration algorithm. The observability aware module selects trajectory sequences that carry enough excitation for the different degrees of freedom required for calibration. Additionally, we provide an analytical study of the effect of noise on the estimated states used for calibration. We also provide a stopping measure which halts the calibration when a satisfactory performance has been achieved.

	Our method offers equivalent performance for rotation estimate and superior outcomes for translation estimate when compared to trajectory alignment method. For future work we will develop a method which can generate a driving route containing observability-aware trajectories which the vehicle can follow to perform online calibration in a minimum amount of time possible.
	
	\section{Acknowledgements}
	This research has been jointly funded by the Swedish Foundation for Strategic Research (SSF) and Scania. The research was also affiliated with Wallenberg AI, Autonomous Systems and Software Program (WASP).
	
	%\addtolength{\textheight}{-12cm}  % This command serves to balance the column
	% lengths on the last page of the document manually. It shortens the textheight
	% of the last page by a suitable amount. This command does not take effect
	%until
	% the next page so it should come on the page before the last. Make sure that
	% you do not shorten the textheight too much.
	
	%%%%%%%%%%%%%%%%%%%%%%%%%%%%%%%%%%%%%%%%%%%%%%%%%%%%%%%%%%%%%%%%%%%%%%%%%%%%%%%%
	
	\bibliographystyle{IEEEtran}
	\bibliography{IEEEabrv,library}
	
	%%Bibtex cleaner: https://flamingtempura.github.io/bibtex-tidy/
	
\end{document}